\documentclass[11pt]{article}

\usepackage[final]{acl}

\usepackage{times}
\usepackage{latexsym}

\usepackage[T1]{fontenc}

\usepackage[utf8]{inputenc}

\usepackage{microtype}

\usepackage{inconsolata}

\usepackage{graphicx}
\usepackage{amssymb}
\usepackage{amsmath}
\usepackage{amsthm}
\usepackage{transparent}
\usepackage{comment}
\usepackage{tikz}
\usepackage{pgfplots}
\usetikzlibrary{external,positioning,shapes,arrows, calc, fit, matrix, arrows.meta, decorations.pathreplacing, shapes.multipart, calligraphy}

\usepackage{booktabs}
\pgfplotsset{compat=1.18}

\title{K-Quantization and its Impact on Output Performance}

\author{Robin Baki Davidsson \\
Lund University \\
  Lund, Sweden \\
  \texttt{robindavidsson@outlook.com} \\\And
  Pierre Nugues \\
Lund University \\
  Lund, Sweden \\
  \texttt{pierre.nugues@cs.lth.se}}

\begin{document}
\maketitle
\begin{abstract}
Recent advancements in large language models (LLMs) have shown their remarkable capacities in many NLP tasks. However, their substantial size often presents challenges for deployment. This necessitates efficient techniques for model compression, with quantization emerging as a prominent solution. Despite its benefits, the exact impact of quantization (from 2- to 6-bit) on the performance and accuracy of LLMs remains an active area of research. 
This paper investigates the performance of eight LLMs at various quantization levels, focusing on tasks such as MMLU-Pro for knowledge processing and reasoning, CRUXEval for code comprehension, and MuSR for reading comprehension. Our results show a consistent trend where higher precision (e.g., 8-bit Q8\_0) yields improved performance, albeit with diminishing returns. Aggressive quantization (e.g., 2-bit Q2\_K) usually retains acceptable accuracy, though some models show a substantial loss in performance. 
Our findings indicate that while lower bit precision generally reduces performance, the impact varies across models and tasks. Larger models show greater resilience to aggressive quantization, but can still undergo significant drops at lower precision levels. Mid-sized models in the 7-9 billion parameter range strike an optimal balance between efficiency and resource usage. Such results provide insights into the trade-offs between model size, quantization, and performance.
\end{abstract}

\section{Introduction}
Large language models (LLMs) are emerging as powerful tools, capable of human-like communication in many areas. However, their significant size presents a challenge when deploying them in real-world applications. This challenge is particularly important in environments where computing resources are limited or data privacy is essential, such as in hospitals, research laboratories, or education. Quantization is a popular way to make these models more compact. It essentially shrinks the models by using less precise numbers for their internal parameters (weights). 

Although quantization helps efficiency, a clearer understanding is needed regarding its impact on model capabilities. More specifically, we would like to know how reducing the precision of the weights affects their ability to reason correctly, understand complex code, or grasp the nuances of long text.

This paper explores the effects of quantization levels on the performance of eight different LLMs, focusing on their reasoning capabilities in knowledge processing tasks, using the MMLU-Pro dataset, code comprehension, using CRUXEval, and understanding long text, using MuSR. Our goal is to find the trade-offs between the efficiency gained from quantization and any potential loss in accuracy or reliability. Our results generally show that performance decreases as quantization gets more aggressive with lower bit precision, though models often remain reasonably accurate even at 2-bit levels. We did observe exceptions nonetheless, where performance dropped sharply at the lowest bit level and the impact depended on the specific model and the task it was performing.

\section{Related Work}
Large language models (LLMs) are facing adoption hurdles due to high computational demand, usage of sensitive data, and strict privacy settings. Quantization can ease these constraints with a minimal sacrifice in performance, and has become an active area of research.

\paragraph{Quantization methods.}

A variety of post-training quantization (PTQ) methods have been developed to compress LLMs without costly retraining. One of the pioneering one-shot methods is GPTQ \citep{frantar2023gptqaccurateposttrainingquantization}, which can accurately compress weights to sizes as low as 3 or 4 bits per parameter. More recently, \citet{lin2024awqactivationawareweightquantization} introduced the activation-aware weight quantization (AWQ), which protects salient weights based on activation magnitudes, often leading to improved performance. Beyond these, libraries such as \href{https://github.com/bitsandbytes-foundation/bitsandbytes}{BitsAndBytes} provide popular on-the-fly quantization implementations for 4- and 8-bit precision, as detailed in \citet{dettmers2022llm} and \citet{dettmers20228bitoptimizersblockwisequantization}. 

\paragraph{Evaluation of quantized models.}
A growing body of research is empirically evaluating the impact of these quantization techniques on model performance, revealing a complex interplay between the quantization technique, bit precision, model architecture, and task type.

Previous notable evaluation studies on quantization methods: \citet{DBLP:conf/aaai/YaoW0YH24} studied GPTQ against older methods. \citet{DBLP:journals/corr/abs-2405-03146} studied the quantization effect using BitsAndBytes. \citet{DBLP:journals/corr/abs-2406-12928} conducted a comprehensive study across a wide range of quantization methods, including GPTQ and AWQ. \citet{DBLP:journals/corr/abs-2406-12928} studied GPTQ, AWQ, SpQR and SmoothQuant. \citep{DBLP:journals/corr/abs-2409-11055} studied GPTQ, AWQ, SmoothQuant, and FP8. Other studies include \citet{DBLP:conf/nips/EgashiraVSHV24}, \citet{xu-etal-2024-beyond-perplexity}, \citet{marchisio-etal-2024-quantization}, \citet{10.5555/3692070.3692558}, \citet{gong-etal-2024-llmc}, and \citet{kurtic-etal-2025-give}.

Recent comprehensive studies establish that high-precision formats are nearly lossless. For instance, \citet{kurtic2025givebf16deathaccuracyperformance} conducted over 500,000 evaluations on the Llama 3.1 family and found that 8-bit quantization is ``effectively lossless'' compared to the BF16 baseline. Their work also showed that 8-bit models exhibit low accuracy degradation (1-3\%), and weight-only 4-bit is often similar to 8-bit performance.

\citet{lee2025exploringtradeoffsquantizationmethods} found that the impact of quantization is also highly dependent on model size. Evaluating models from 1B to 405B parameters found that smaller models can suffer severe accuracy drops at 4-bit precision, while larger models (e.g., 70B scale) maintain much more stable performance.

\paragraph{Contribution of this work.}
Prior work such as \citet{jin2024comprehensiveevaluationquantizationstrategies} has evaluated methods like GPTQ \citep{frantar2023gptqaccurateposttrainingquantization} and SpQR \citep{dettmers2023spqr}, and other comprehensive studies have compared AWQ and GPTQ across various model sizes. However, a focused investigation into the effects of k-quantization remains less explored. Our work contributes to this landscape by extending the analysis to four different modern model families (Llama, Gemma, Phi, and Mistral) and evaluating a wide range of low-bit precisions (from 2-bit to 6-bit) using the k-quant technique implemented in llama.cpp \citep{githubHubggufk, ibmgranitequantization}. 


\section{Background}

This section describes the underlying components behind large language models. We focus primarily on the weights, which are directly affected by the compression effect from quantization. We also describe how the model output probability can be utilized and represented as a function of the weights.

\subsection{Autoregressive Models}\label{sec:autoreg}

Autoregressive models are probabilistic models designed specifically for sequence prediction. Autoregressive models form the fundamental basis for text generation in most modern LLMs. 

\paragraph{Definition.}

The probability of the next symbol in a sequence, $x_i$, is given by:
\begin{equation}\label{eq:prob_f}
    p(x_i | x_{1}, ..., x_{i-2}, x_{i-1}),
\end{equation}
where $x_{1}, ..., x_{i-2}, x_{i-1}$ denotes the sequence of past symbols. Essentially, these models are adjusted to map the context into a probability distribution over all potential future values.  


\paragraph{Perplexity.}
Building upon the predictive capabilities of autoregressive models, perplexity is a critical evaluation metric  \citep{zhong2018affectrichneuralconversationalmodel}. Perplexity quantifies how effectively a model predicts a word sequence, and hence text and corpora. It does this by calculating the uncertainty associated with the predictions given the preceding context \citep{kuribayashi-etal-2021-lower}.

A lower perplexity indicates that the model more accurately anticipates the next token in a sequence, signifying stronger language understanding and generation abilities \citep{zhong2018affectrichneuralconversationalmodel}. For a sequence of $N$ tokens, perplexity is expressed as:

\begin{equation}\label{eq:ppl}
    \text{PPL} = \prod_{i=0}^N p(w_i | w_{<i})^{-\frac{1}{N}}
\end{equation}
Eq. \ref{eq:ppl} is derived from entropy in information theory \citep{kuribayashi-etal-2021-lower, shannon1948mathematical,shannon1948mathematical2}. 

\paragraph{Context window.}
The context window is the count of preceding symbols utilized by the model for making future predictions \citep{Dunn_2023}. To carry out the calculation, we process text sequentially within a fixed-size context window. When the context exceeds the predefined limit (maximum tokens for a specific model), the window is shifted further along the text.

\subsection{Weights}
The size of a model specifically refers to its number of parameters or weights. Notably, `open-weight models' are those for which the weights have been made publicly available, allowing for local usage without dependency on an online service.

During the training process, the weights are iteratively updated to find the optimal function that fits the data. The weights are adjusted to minimize the difference between the model's predicted output and the target output based on a predetermined loss and optimization algorithm. The adjustment process is responsible for refining the weights. These optimized weights are effective in capturing the underlying structure of the dataset, allowing the LLM to generate text or code that is both grammatically correct and semantically meaningful.

We can rewrite Eq. \ref{eq:prob_f} using weights as:
\begin{equation}\label{eq:func}
    \hat{x}_i = f(x_{1}, ..., x_{i-2}, x_{i-1}; W),
\end{equation}
where $f$  takes the past elements $x_{1}, ..., x_{i-2}, x_{i-1}$ as input and makes the prediction $\hat{x}_i$, and where $W$ represents the weights.

\subsection{Weights in Model Structures}
Most autoregressive models build on the transformer architecture \citep{vaswani2023attentionneed}, restricted to its decoder part. The number of parameters is roughly a function of the number of layers in the decoder. Nonetheless, individual decoder components are rapidly evolving with the emergence of newer activation functions and positional encoding, deviating from the original architecture.

\paragraph{Weight representations.} Weights of large language models are typically stored as floating-point numbers. A common format is bfloat16, which uses 16 bits per number. Bfloat16 prioritizes the representation of a wider range of values at the cost of slightly lower precision compared to traditional float16. Figure~\ref{fig:float16} shows the bit distribution between the mantissa and the exponent of float16 and Figure~\ref{fig:bfloat16}, the bit distribution of bfloat16.

\begin{figure}[b]
\vspace{-0.5cm}
\centering
    \setlength{\unitlength}{0.5mm} 
    \hfill
    \begin{picture}(150, 30)

        \put(0, 10){\line(1, 0){8}}  
        \put(0, 20){\line(1, 0){8}}  
        \put(0, 10){\line(0, 1){10}} 
        \put(8, 10){\line(0, 1){10}} 
        
        \put(12, 10){\line(1, 0){40}}  
        \put(12, 20){\line(1, 0){40}}  
        \put(12, 10){\line(0, 1){10}}  
        \put(52, 10){\line(0, 1){10}}  
        
        \put(56, 10){\line(1, 0){80}}  
        \put(56, 20){\line(1, 0){80}}  
        \put(56, 10){\line(0, 1){10}}  
        \put(136, 10){\line(0, 1){10}} 
        
        \put(4, 24){\makebox(0,0){\scriptsize 0}}
        
        \multiput(12, 10)(8, 0){5}{\line(0, 1){10}} 
        \put(16, 24){\makebox(0,0){\scriptsize 1}}
        \put(24, 24){\makebox(0,0){\scriptsize 2}}
        \put(32, 24){\makebox(0,0){\scriptsize 3}}
        \put(40, 24){\makebox(0,0){\scriptsize 4}}
        \put(48, 24){\makebox(0,0){\scriptsize 5}}
        
        \multiput(56, 10)(8, 0){10}{\line(0, 1){10}} 
        \put(60, 24){\makebox(0,0){\scriptsize 6}}
        \put(68, 24){\makebox(0,0){\scriptsize 7}}
        \put(76, 24){\makebox(0,0){\scriptsize 8}}
        \put(84, 24){\makebox(0,0){\scriptsize 9}}
        \put(92, 24){\makebox(0,0){\scriptsize 10}}
        \put(100, 24){\makebox(0,0){\scriptsize 11}}
        \put(108, 24){\makebox(0,0){\scriptsize 12}}
        \put(116, 24){\makebox(0,0){\scriptsize 13}}
        \put(124, 24){\makebox(0,0){\scriptsize 14}}
        \put(132, 24){\makebox(0,0){\scriptsize 15}}
        
        \put(4, 5){\makebox(0,0){\textbf{Sign}}}
        \put(32, 5){\makebox(0,0){\textbf{Exponent}}}
        \put(96, 5){\makebox(0,0){\textbf{Mantissa}}}
        
        \multiput(8, 10)(0, 2){5}{\line(0, 1){1}}  
        \multiput(52, 10)(0, 2){5}{\line(0, 1){1}} 
        
    \end{picture}
    \caption{Distribution of the 16 bits in float16.}
    \label{fig:float16}
    \hfill
    \begin{picture}(150, 30)

        \put(0, 10){\line(1, 0){8}}  
        \put(0, 20){\line(1, 0){8}}  
        \put(0, 10){\line(0, 1){10}} 
        \put(8, 10){\line(0, 1){10}} 
        
        \put(12, 10){\line(1, 0){64}}  
        \put(12, 20){\line(1, 0){64}}  
        \put(12, 10){\line(0, 1){10}}  
        \put(76, 10){\line(0, 1){10}}  
        
        \put(80, 10){\line(1, 0){56}}  
        \put(80, 20){\line(1, 0){56}}  
        \put(80, 10){\line(0, 1){10}}  
        \put(136, 10){\line(0, 1){10}} 
        
        \put(4, 24){\makebox(0,0){\scriptsize 0}}
        
        \multiput(12, 10)(8, 0){8}{\line(0, 1){10}} 
        \put(16, 24){\makebox(0,0){\scriptsize 1}}
        \put(24, 24){\makebox(0,0){\scriptsize 2}}
        \put(32, 24){\makebox(0,0){\scriptsize 3}}
        \put(40, 24){\makebox(0,0){\scriptsize 4}}
        \put(48, 24){\makebox(0,0){\scriptsize 5}}
        \put(56, 24){\makebox(0,0){\scriptsize 6}}
        \put(64, 24){\makebox(0,0){\scriptsize 7}}
        \put(72, 24){\makebox(0,0){\scriptsize 8}}
        
        \multiput(80, 10)(8, 0){7}{\line(0, 1){ 10}} 
        \put(84, 24){\makebox(0,0){\scriptsize 9}}
        \put(92, 24){\makebox(0,0){\scriptsize 10}}
        \put(100, 24){\makebox(0,0){\scriptsize 11}}
        \put(108, 24){\makebox(0,0){\scriptsize 12}}
        \put(116, 24){\makebox(0,0){\scriptsize 13}}
        \put(124, 24){\makebox(0,0){\scriptsize 14}}
        \put(132, 24){\makebox(0,0){\scriptsize 15}}
        
        \put(4, 5){\makebox(0,0){\textbf{Sign}}}
        \put(44, 5){\makebox(0,0){\textbf{Exponent}}}
        \put(108, 5){\makebox(0,0){\textbf{Mantissa}}}
        
        \multiput(8, 10)(0, 0){5}{\line(0, 1){1}}  
        \multiput(76, 10)(0, 0){5}{\line(0, 1){1}} 
        
    \end{picture}
    \caption{Distribution of the 16 bits in bfloat16.}
    \label{fig:bfloat16}
\end{figure}

\paragraph{Weight quantization.}
Quantization emerges as a crucial technique for reducing the memory footprint and accelerating the computation of LLMs, making them suitable for deployment on resource-constrained devices \citep{gong2024makesquantizationlargelanguage}. This process is done by converting the model weights so that they use fewer bits per value, thus reducing precision. Lower precision weights require less storage, which allows for faster data transfer and arithmetic operations. 

This paper utilizes post-training quantization (PTQ), a method applied after the model is trained to avoid the need for costly retraining \citep{frantar2023gptqaccurateposttrainingquantization, huang2024billmpushinglimitposttraining}. A prevalent PTQ strategy is affine block quantization \citep{HubdocsQuantization}. The method will utilize double quantization \citep{dettmers2023qloraefficientfinetuningquantized}; it applies quantization to the quantization constants. 

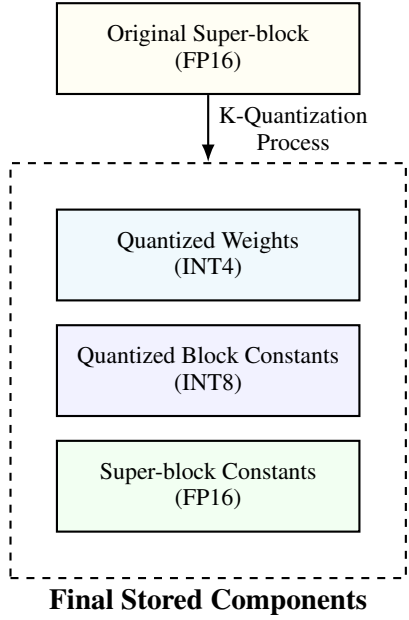
\begin{figure}[tb]
\centering
\begin{tikzpicture}[
    node distance=0.3cm and 0.5cm,
    block/.style={draw, thick, minimum width=4cm, minimum height=1.2cm, align=center},
    grid_block/.style={
        block,
        path picture={
            \draw[step=3mm, black!20] (path picture bounding box.south west) grid (path picture bounding box.north east);
        }
    },
    meta_block/.style={
        block, fill=black!20, font=\small
    },
    arrow/.style={-Latex, thick, draw=black}
]

\node[meta_block, fill=yellow!5] (W_orig) {Original Super-block \\ (FP16)};

\node[meta_block, fill=cyan!5, below=1.5cm of W_orig] (W_quant) {Quantized Weights \\ (INT4)};
\node[meta_block, fill=blue!5, below=of W_quant] (C_quant) {Quantized Block Constants \\ (INT8)};
\node[meta_block, fill=green!5, below=of C_quant] (C_super) {Super-block Constants \\ (FP16)};

\node[draw, thick, dashed, inner sep=6mm, fit=(W_quant) (C_quant) (C_super),
      label={[font=\bfseries]below:Final Stored Components}] (final_box) {};

\draw[arrow] (W_orig.south) -- (final_box.north)
    node[midway, right, font=\small, align=center] {K-Quantization \\ Process};

\end{tikzpicture}
\caption{The structure of a K-quantized super-block. The process decomposes the original high-precision block into three final components.}
\label{fig:kquant_structure}
\end{figure}

We chose the $k$-quantization method, as implemented in \verb=llama.cpp= for this study. K-quantization utilizes a hierarchical structure \citep{githubgguf-docs}, as seen in figure \ref{fig:kquant_structure}. The method groups a set of the model’s high-precision weights into smaller fixed-size blocks, and for each block, a high-precision scalar ($S_k$) and offset ($\alpha_k$) are calculated. Those are stored in a larger structure called a super-block.

The constants for each block are no longer stored with high precision; they are quantized to INT8. Another set of high-precision constants is then stored for the entire super-block. These super-block constants are used to de-quantize the block-level constants during inference \citep{githubgguf-docs}. The quantized weights are mapped to INT4 bins and stored as such. K-quantization employs a mixed-precision strategy for the weights to balance model size and performance. The method does not apply the same bit quantization to all weights; some are more sensitive to quantization error and are assigned higher precision.

\paragraph{Llama.cpp and GGUF.}
For model evaluation and execution, we utilized the \verb=llama.cpp= software library \citep{githubGitHubGgerganovllamacpp} with its Python bindings \citep{llama-cpp-python}. This library stores the quantized models in a binary file format called GGUF. GGUF contains the tensor data and metadata. It supports the 2- to 8-bit quantization types that we describe in this paper \citep{githubGitHubGgerganovllamacpp, githubHubdocsdocshubggufmdMain}.

\section{Models}\label{sec:models}

\paragraph{Llama.} Llama 3 models \citep{meta:ai} mark a significant advancement in the field of open-weight LLMs. The models are available in two sizes: 8B and 70B. They are designed to be run locally and provide tools for exploring the capabilities of large language models. Smaller models, like the 8B version, can be deployed on less powerful hardware, making them more accessible for individual users or smaller organizations. The larger 70B model, while requiring significant computational power, offers enhanced performance and capabilities suitable for tasks demanding higher accuracy and complexity. 

Table \ref{tab:datatypes} shows the quantized data types that we used, along with their corresponding file sizes. In addition, the table presents the settings that we used in the evaluation process.

\begin{table*}[tb]
\centering
\resizebox{\textwidth}{!}{
\begin{tabular}{lrrrrrrrrr}
\hline
\textbf{Model}&\textbf{\# Parameters}&\textbf{Context length}&\textbf{Max token out}&\textbf{Q8\_0}&\textbf{Q6\_K}&\textbf{Q5\_K} &\textbf{Q4\_K}&\textbf{Q3\_K}&\textbf{Q2\_K}\\
\hline
LLama 3 Instruct&8&4096&2048&8.5& 6.6& --&5.0& 4.0& 3.2\\
LLama 3 Instruct& 70 & 4096 & 2048&75&--& --&42.5& --&26.4\\
Gemma v1.1 Instruct & 2 & 4096 & 2048 & 2.7& 2.1& 1.8&1.6& 1.4& 1.2\\
Gemma 2 Instruct & 9 & 4096 & 2048 &9.8& 7.6&--& 5.8&4.8& 3.8\\
{Gemma 2 Instruct}& 27 & 4096 & 2048 &28.9& --&19.4 & --&13.4& 10.4\\
Phi-3 Mini Instruct & 4 & 4096 & 2048 &4.1& 3.1& --&2.4&2.0& 1.4\\
Phi-3 Medium Instruct & 14 & 4096 & 2048 &14.8& --&10.1&--&6.9& 5.1\\
{Mistral Instruct v0.3}
& 7 & 4096 & 2048 &7.7& 5.9& --&4.4&3.5&2.7\\
\hline
\end{tabular}}
\caption{Models, number of parameters in billions, weight data types and their model sizes in gigabytes.}
\label{tab:datatypes}
\end{table*}


\paragraph{Gemma.}
Gemma 2B \citep{gemma:Banks_Warkentin_2024} was first released as a state-of-the-art model in its size bracket. Later, \citet{gemma2:Farabet_Warkentin_2024} expanded the Gemma family with two larger models: Gemma 2 9B and Gemma 2 27B, see Table \ref{tab:datatypes}. We selected these more recent models from the same family so that we could compare how the different model sizes affect performance under various conditions. This should provide us with insights on their scalability and effectiveness.

\paragraph{Phi.}
Phi-3 is another family of models \citep{phi3:Bilenko_2024}, which comes in two sizes: Phi-3-small with 7B parameters and Phi-3-medium with 14B, see Table \ref{tab:datatypes}. According to \citet{phi3:Bilenko_2024}, these models represent the state-of-the-art for small models within their respective size brackets. We chose it to determine the quantization effect on small models.

\paragraph{Mistral.}
Mistral 7B \citep{Mistral:AIteam} has become a benchmark for high-performance open-weight models. It is frequently selected for tasks and research involving smaller models due to its size and output quality. We added it as it has an intermediate size between Phi, Llama, and Gemma.

\section{Datasets}
We applied the LLM evaluation to the  MMLU-Pro, CRUXEval, and MuSR datasets. They represent three different use cases: MMLU-Pro focuses on the model's knowledge capability, CRUXEval evaluates the model's ability to understand code and the logic behind the code, and MuSR assesses the model's capability to understand long context texts and be coherent.

The three datasets have been released after the cutoff date of the data used by the LLMs. This separation is intentional. It minimizes the chance that these datasets were accidentally included in the training dataset of the LLMs. This is essential for maintaining the integrity of the evaluation process. It ensures the models are tested on unseen data.

\paragraph{MMLU-Pro.}
The Massive Multitask Language Understanding Pro dataset (MMLU-Pro) \citep{wang2024mmluprorobustchallengingmultitask} consists of questions and answers divided into 14 categories. Each question in this dataset is structured with a chain-of-thought (CoT) explanation and ten possible answer choices, promoting a more comprehensive evaluation of the model's cognitive processes. It is designed with two purposes: assess LLMs in both knowledge and reasoning capabilities, and to reduce the probability of models arbitrarily guessing correct answers \citep{wang2024mmluprorobustchallengingmultitask}. 

We employed a subset of the MMLU-Pro dataset due to computational and time constraints. We utilized the first 100 questions from each of the 14 categories, resulting in a total of 1400 questions. While this subset is not exhaustive, it still provides a robust and diverse sample for evaluating the LLM's performance across multiple domains.

\paragraph{CRUXEval.}

The utilization of LLMs in scientific domains has experienced a significant surge, particularly for code generation \citep{nejjar2024llmsscienceusagecode}. To address this growing application, CRUXEval (Code Reasoning, Understanding, and eXecution Evaluation) was specifically developed to assess code comprehension and reasoning capabilities of LLMs \citep{gu2024cruxevalbenchmarkcodereasoning}. CRUXEval comprises 800 Python functions, each accompanied by corresponding input-output pairs. It will enable us to evaluate the LLM's ability to understand code, process inputs, and predict accurate outputs.

\paragraph{MuSR.}
The MuSR (Multistep Soft Reasoning) dataset was designed to measure the reasoning capabilities of LLMs in complex scenarios without relying on CoT prompting \citep{sprague2024musrtestinglimitschainofthought}. The dataset is split into three parts: object placements, team allocation, and murder mysteries. This evaluation technique contains narrative texts of approximately 1000 words each, accompanied by related questions and multiple-choice answers.

\paragraph{Wikitext.}
For the assessment of model perplexities, we utilized the \verb=wikitext-2-raw-v1= dataset \citep{merity2016pointer}. Two primary considerations motivated our choice. It comprises clean, verified texts extracted from Wikipedia articles, ensuring high quality and well-structured content. This is further amplified by a diverse and extensive vocabulary range, as well as the coverage of various topics and writing styles. It is also compatible with the llama.cpp library and is one of its recommended datasets \citep{githubGitHubGgerganovllamacppperplexity}. 
We specifically used the `test' partition of the dataset.

We acknowledge that the selected datasets primarily evaluate short-form or multiple-choice answers. As noted by \citet{wang2024myanswercfirsttoken}, such benchmarks may not fully capture the generative and long-form reasoning capabilities of LLMs. However, they provide a standardized and reproducible framework for assessing core knowledge and reasoning under quantization.

\section{Methodology}
The evaluation methodology focused on analyzing the response structure and information accuracy. We assessed the models' ability to understand complex queries and generate answers that precisely follow the provided instructions. This approach aims to determine how well LLMs interpret questions and produce relevant responses according to the instructions.

\subsection{Approach}

\paragraph{Dataset accuracy.}
In our evaluation of LLMs, accuracy on the dataset refers to the proportion of correct predictions made by the model relative to the total number of questions. The mean accuracy is the geometric mean across the tasks. A higher mean accuracy score indicates a greater precision in handling a variety of tasks and queries. Furthermore, we used the mean accuracy as a way to compute the resource efficiency. We calculated it by dividing the mean accuracy of the model by the size in gigabytes of the model. 

\paragraph{Few-shot prompting.}\label{subsec:few-shot}
Understanding the cognitive process is often necessary to apply logical reasoning when solving a task. This is essentially the purpose of few-shot prompting \citep{brown2020languagemodelsfewshotlearners}. In this context, we provide the LLM with one or more examples of how previous tasks have been completed along with comprehensive guidelines. Subsequently, we give the LLM the task to undertake a similar assignment. This approach has demonstrated increased performance across various datasets.

\paragraph{Chain-of-thought (CoT).}
Multiple reasoning steps pose a significant challenge for LLMs, often leading to inaccurate responses, including instances of hallucination \citep{tonmoy2024comprehensivesurveyhallucinationmitigation}. To address this issue, \citet{wei2023chainofthoughtpromptingelicitsreasoning} introduced the chain-of-thought (CoT) prompting technique. CoT's structured approach provides examples in the previous few-shot prompts. It encourages LLMs to engage in a methodical, step-by-step reasoning process when generating responses. \citet{wei2023chainofthoughtpromptingelicitsreasoning} showed it significantly enhanced performance across various reasoning tasks. It facilitates more thorough and logical deliberation in the formulation of answers. We used CoT only with the datasets that contained this feature. 

\paragraph{8-bit as baseline.}
For this study, we use the 8-bit quantized model (Q8\_0) as baseline. We acknowledge that the standard industry practice is to compare against full-precision models. However, due to computational constraints, we limited our baseline to 8-bit. As such, our results reflect the drop from an already quantized state. The true performance loss from full precision may be slightly larger. \citet{dettmers20228bitoptimizersblockwisequantization} and \citet{jin2024comprehensiveevaluationquantizationstrategies} showed that 8-bit quantization can achieve performance figures close to those obtained with full precision across a wide range of tasks and model architectures. These findings suggest that an 8-bit representation preserves enough information for task evaluation without a significant loss of precision.


\subsection{Implementation}
The experimental setup explores the influence of model sizes, quantization techniques, and datasets on performance. This provides a comprehensive evaluation of LLM performance across various task types. We utilized the llama.cpp library to deploy and create $k$-quants of the large language models.

\paragraph{Perplexity tool.}
The Llama.cpp\footnote{https://github.com/ggerganov/llama.cpp} library provides tools for calculating perplexity \citep{githubGitHubGgerganovllamacppperplexity,githubGitHubGgerganovllamacppDiscussion}. \citet{gguf:JohannesGaessler} gives additional details on the usage of llama.cpp and the metrics it provides. His work made the documentation on tool utilization and its importance much easier to understand.

\paragraph{Settings.} \label{subsec:settings}
Table \ref{tab:prompts} shows the few-shot settings for Llama 3, Gemma, Gemma 2, Phi 3, and Mistral. We used the CoT instructions when they were part of the dataset. Table \ref{tab:datatypes} shows the quantization levels we assessed. Due to resource and time constraints, we had to limit the number of models with a higher parameter count. This explains why we tested fewer quantization levels for these models. 

\begin{table}[tb]
\centering
\resizebox{\columnwidth}{!}{
\begin{tabular}{lr|rrr}
\hline
\textbf{Model}&\textbf{\# P.}&\textbf{MMLU-Pro}&\textbf{CRUXEval}&\textbf{MuSR}\\
\hline
LLama 3 Instruct&8&4&1&1\\
LLama 3 Instruct& 70 & 1&1&0\\
Gemma v1.1 Instruct & 2&4&1&1 \\
Gemma 2 Instruct & 9&4&1&1 \\
{Gemma 2 Instruct}& 27&1&1&0\\
Phi-3 Mini Instruct & 4&1&1&0 \\
Phi-3 Medium Instruct & 14 &4&1&0\\
{Mistral Instruct v0.3}
& 7&1&1&0 \\
\hline
\end{tabular}}
\caption{Models and number of shots per prompt}
\label{tab:prompts}
\end{table}

\section{Evaluation Results}
We organized the results by their respective architectures, showing both the perplexity and accuracy scores. This setup makes it easy to compare how they perform at their various levels. Note that the goal of this work is only to study how weight quantization affects the individual models. The results of each model should not be compared to those of the other models.

\subsection{The Gemma Family}
The Gemma 2B v1.1 model obtained perplexity scores ranging from 30.02 to 39.86 across quantization levels (Table \ref{tab:gemma-2b}). Notably, the Q8\_0 quantization achieved the lowest perplexity, indicating superior language modeling performance. The accuracy scores for this model varied across datasets, with MMLU-Pro scores ranging from 11.92\% to 15.42\%, MuSR scores between 37.52\% and 40.69\%, and CRUXEval scores from 22.72\% to 26.72\% (Table \ref{tab:gemma-2b}). The mean accuracy across all datasets peaked at 27.26\% for the Q5\_K quantization.

\paragraph{Gemma 2B v1.1.}

In Table \ref{tab:gemma-2b}, the perplexity values are presented along with their margins of error for each quantization level, which range from scores of 30.02 to 39.86. This trend suggests that higher precision quantization preserves more of the model's predictive capabilities.

\begin{table*}[tb]
\begin{center}
\resizebox{\textwidth}{!}{
\begin{tabular}{ll||r|rrrrr}
\toprule
\textbf{Model}&\textbf{Quantization} &\textbf{Perplexity}& \textbf{MMLU-Pro} & \textbf{MuSR} & \textbf{CRUXEval} & \textbf{Geometric mean} & \begin{tabular}[c]{@{}l@{}}\textbf{Mean / GB}\end{tabular} \\
    \midrule 
Gemma 2 2B&Q2\_K & 39.86 $\pm$ 0.43& 11.92 & 37.52 & 22.72 & {21.66} & \textbf{18.1} \\
&Q3\_K & 32.11 $\pm$ 0.35& 13.78 & 39.89 & 26.47 & {24.41} & 17.4 \\
&Q4\_K & 30.91 $\pm$ 0.34& 14.49 & 40.29 & \textbf{26.72} & {24.99} & 15.6 \\
&Q5\_K & 30.25 $\pm$ 0.33& 15.13 & \textbf{40.69} & 25.97 & {25.19} & 14.0 \\
&Q6\_K & 30.16 $\pm$ 0.33& \textbf{15.42} & 40.16 & 25.97 & \textbf{25.24} & 12.0 \\
&Q8\_0 & \textbf{30.02 $\pm$ 0.33}& 14.78 & 40.55 & 25.22 & 24.73 & 9.2 \\
    \bottomrule
    \midrule
Gemma 2 9B&Q2\_K & 10.16$\pm$ 0.07& 39.54 & 42.01 & 37.95 & 39.80 & \textbf{10.5}\\
&Q3\_K & 9.16 $\pm$ 0.07& 43.97 & \textbf{44.12} & 37.83 & 41.87 & 8.7 \\
&Q4\_K & 8.92 $\pm$ 0.07& \textbf{45.61} & \textbf{44.12} & 36.45 & 41.86 & 7.2 \\
&Q6\_K & 8.83 $\pm$ 0.07& 44.25 & 43.33 & \textbf{38.45} & \textbf{41.93} & 5.5\\
&Q8\_0 & \textbf{8.82 $\pm$ 0.07}& 45.04 & 43.20 & 37.33 & 41.72 & 4.3 \\
\bottomrule
\midrule
Gemma 2 27B&Q2\_K & 9.17 $\pm$ 0.06& 42.90 & 42.54 & 43.32 & 42.92 & \textbf{4.1} \\
&Q3\_K & 7.76 $\pm$ 0.05& 48.11 & \textbf{45.31} & 53.31 & 48.80 & 3.6 \\
&Q5\_K & 7.25 $\pm$ 0.05& 46.11 & 45.44 & \textbf{54.56} & 48.53 & 2.5 \\
&Q8\_0  & \textbf{7.19 $\pm$ 0.05}& \textbf{49.86} & 44.78 & 54.18 & \textbf{49.46} & 1.7 \\
\bottomrule
    \midrule 
Llama 3 8B
&Q2\_K & 11.36 $\pm$ 0.08& 17.84 & \textbf{40.03} & 25.72 & 26.38 & 8.2 \\
&Q3\_K & 9.02 $\pm$ 0.07& 32.62 & 39.37 & 35.96 & 35.88 & \textbf{9.0} \\
&Q4\_K & 8.49 $\pm$ 0.06& 32.33 & 36.20 & \textbf{36.95} & 35.10 & 7.0 \\
&Q6\_K & \textbf{8.36 $\pm$ 0.06}& 32.69 & 37.91 &37.70 & \textbf{36.02} & 5.5 \\
&Q8\_0 & \textbf{8.36 $\pm$ 0.06}& \textbf{38.31} & 37.08 & 31.55 & 35.52 & 4.2 \\
    \bottomrule
    \midrule
Llama 3 70B
&Q2\_K & 6.87 $\pm$ 0.05 & 38.83 & 41.08 & 46.19 & 41.92 & \textbf{1.6}\\
&Q4\_K & 5.33 $\pm$ 0.04& 53.53 & \textbf{43.86} & 51.69 & 49.51 & 1.2 \\
&Q8\_0 & \textbf{5.18 $\pm$ 0.03}& \textbf{55.32} & 43.46 & \textbf{55.43} & \textbf{51.08} & 0.7 \\
\bottomrule
    \midrule 
Phi 3 mini
&Q2\_K & 195.79 $\pm$ 1.69& 6.14 & 7.66 & 0.37 & 2.59 & 1.9 \\
&Q3\_K & 7.23 $\pm$ 0.04& 20.56 & 5.81 & \textbf{13.61}& \textbf{11.76} &\textbf{5.9}\\
&Q4\_K & 6.69 $\pm$ 0.04& 21.48 & 7.00 & 9.49 & 11.26  & 4.7 \\
&Q6\_K & 6.42 $\pm$ 0.04& \textbf{27.69} & 6.74 & 6.49 & 10.66  & 3.4 \\
&Q8\_0 & \textbf{6.41 $\pm$ 0.04}& 27.48 & \textbf{7.74} & 6.61 & 11.20  & 2.7 \\
    \bottomrule
    \midrule
Phi 3 medium
&Q2\_K & 57.07 $\pm$ 0.44& 3.21 & 0.26 & 0.62 & 0.80 & 0.2\\
&Q3\_K & 5.43 $\pm$ 0.03& 44.54 & \textbf{27.61} & 28.09 & \textbf{32.57} &\textbf{4.1} \\
&Q5\_K & 4.65 $\pm$ 0.03& \textbf{47.39} & 24.04 & \textbf{28.34} & 31.84 &3.2 \\
&Q8\_0 & \textbf{4.57 $\pm$ 0.03}& 47.11 & 25.76 & 26.72 & 31.89 &2.2 \\
\bottomrule
    \midrule 
Mistral 7B v0.3
&Q2\_K & 6.98 $\pm$ 0.04& 24.05 & 40.56 & \textbf{19.60} & {26.74} & \textbf{9.9} \\
&Q3\_K & 6.32 $\pm$ 0.04& 29.84 & 41.74 & 18.73 & \textbf{28.57} & 8.2 \\
&Q4\_K & 6.21 $\pm$ 0.04& 30.69 & \textbf{43.46} & 10.74 & 24.29 & 5.5 \\
&Q6\_K & 6.19 $\pm$ 0.04& 30.69 & 40.95 & 13.73 & 25.84 & 4.4 \\
&Q8\_0 & \textbf{6.18 $\pm$ 0.04}& \textbf{31.41} & 41.08 & 12.48 & 25.25 & 3.3  \\
\bottomrule
\end{tabular}}
\end{center}
\caption{Perplexity of LLM families for each quantization with margins of error. Lower is better. Comparison of performance metrics (in percentage) for  multiple datasets and their geometric mean values.}
\label{tab:gemma-2b}
\end{table*}

All of the accuracy scores for this model varied across the datasets, with MMLU-Pro scores ranging from 11.92\% to 15.42\%, MuSR scores between 37.52\% and 40.69\%, and CRUXEval scores from 22.72\% to 26.72\% (Table \ref{tab:gemma-2b}). Interestingly, the mean accuracy across all datasets peaked at 25.24\% for the Q5\_K quantization, a slight improvement over Q8\_0. This suggests that there might be an optimal quantization, where the model maintains most of its performance while significantly reducing its size. The last column shows the performance efficiency of the model, where we divide the mean by the number of gigabytes of the model. It decreases from 18.1 for Q2\_K to 9.2 for Q8\_0.


\paragraph{Gemma 2 9B.}
The Gemma 2 9B model showed strong performance, with perplexity scores ranging from 8.82 to 10.16 (Table \ref{tab:gemma-2b}). It showed consistent performance across the quantization levels, with mean accuracy scores ranging from 39.8\% to 41.93\% (Table \ref{tab:gemma-2b}). The MMLU-Pro dataset saw particularly notable changes at lower levels.

This model showed more consistent performance, with mean accuracy scores ranging from 39.83\% to 41.93\% (Table \ref{tab:gemma-2b}). The MMLU-Pro dataset saw scores reaching up to 45.61\% for the Q4\_K quantization. This suggests that the larger model size allows for better retention of knowledge and reasoning at lower quantization levels. The performance efficiency of the model decreases from 10.5 for Q2\_K to 4.3 for Q8\_0 (Table \ref{tab:gemma-2b}), last column).


\paragraph{Gemma 2 27B.}
Gemma 2 27B, demonstrated the best performance in the Gemma family, with perplexity score ranges from 7.19 to 9.17 (Table \ref{tab:gemma-2b}).

Table \ref{tab:gemma-2b} shows all the evaluations across all the datasets, with mean scores ranging from 42.92\% to 48.91\%. The MMLU-Pro dataset saw scores up to 48.11\%, while the CRUXEval dataset reached 54.56\% for the Q5\_K quantization. This performance gap shows the impact of model size on task-specific capabilities. The last column shows the performance efficiency of the model decreasing from 4.1 for Q2\_K to 1.7 for Q8\_0.

\subsection{The Llama 3 Family}
The Llama 3 family showed resilience against the impact of quantization across the model sizes, demonstrating the robustness of its architecture.

\paragraph{Llama 3 8B.}   
The Llama 3 8B model showed competitive performance with perplexity scores ranging from 8.36 to 11.36 (Table \ref{tab:gemma-2b}). The accuracy scores for this model varied significantly across quantization levels, with a notable drop in performance for the Q2\_K quantization.

The accuracy scores for Llama 3 8B did not vary much across most of the quantization levels, with a notable drop in performance for the Q2\_K quantization. Mean accuracy scores ranged from 26.38\% to 36.02\% (Table \ref{tab:gemma-2b}). The Q6\_K quantization achieved the highest overall performance. Q3\_K scored close to the highest mean score, suggesting that it might offer the optimal balance between model size reduction and performance retention. The performance efficiency of the model decreased from 9.0 for Q3\_K to 4.2 for Q8\_0 (Table \ref{tab:gemma-2b}, last column).

\paragraph{Llama 3 70B.}
Llama 3 70B showed strong performance across the various quantization levels, with perplexity scores ranging from 5.18 to 6.87 (Table \ref{tab:gemma-2b}). The Q8\_0 quantization achieved the lowest perplexity of 5.18. 

Table \ref{tab:gemma-2b} shows all the accuracy evaluations across all the datasets. The accuracy scores showed improvement with higher quantization levels, with mean scores ranging from 41.92\% to 51.08\% (Table \ref{tab:gemma-2b}). The CRUXEval dataset reached 55.43\% accuracy for the Q8\_0 quantization, demonstrating the model's strong code comprehension. The consistency in MuSR scores across quantization levels (ranging from 41.08\% to 43.86\%) suggests that the model's text comprehension capabilities are quite resilient to quantization effects. The MMLU-Pro dataset saw scores up to 55.32\% for Q8\_0 and dropping to 38.83\% for Q2\_K, indicating a significant loss of knowledge. The performance efficiency of the model decreased from 1.6 for Q2\_K to 0.7 for Q8\_0.

\subsection{The Phi 3 Family}
The Phi 3 family shows interesting results, particularly on the performance of 2-bit quantization.

\paragraph{Phi 3 Mini 4B.} 
The Phi 3 Mini model showed the widest range of perplexity scores, from 6.41 to an unusually high 195.79 for the Q2\_K quantization (Table \ref{tab:gemma-2b}). This extreme variation was reflected in the accuracy scores, where the Q2\_K quantization performed poorly across all datasets, achieving a mean accuracy of only 2.59\% (Table \ref{tab:gemma-2b}). However, the other quantization levels performed more consistently, with mean accuracy ranging from 10.66\% to 11.76\%. The performance efficiency of the model decreased from 6.7 for Q3\_K to 3.4 for Q8\_0 (Table \ref{tab:gemma-2b}, last column).

\paragraph{Phi 3 Medium 14B.}    

The Phi 3 Medium 14B model demonstrated the same issue at lower quantization. Perplexity scores ranged from 4.57 to 57.07 (Table \ref{tab:gemma-2b}), with the Q2\_K quantization again showing significantly worse performance. Accuracy scores for this model were more consistent across the Q3\_K, Q5\_K, and Q8\_K quantizations, with mean accuracies ranging from 31.84\% to 32.57\%. However, Q2\_K saw a drastic loss (Table \ref{tab:gemma-2b}). The MMLU-Pro dataset saw particularly strong performance, with scores reaching up to 47.39\% for the Q5\_K quantization. This suggests that the Phi 3 Medium model is capable of maintaining its performance even at lower quantization levels, provided they are not too aggressive. The performance efficiency of the model decreased from 4.1 for Q3\_K to 2.2 for Q8\_0 (Table \ref{tab:gemma-2b}, last column). A notable result is the score of Q2\_K which is as low as 0.2.

\subsection{Mistral 7B v0.3.}    

The Mistral 7B v0.3 model demonstrates a clear increase in the perplexity values as the model is further quantized. However, this increase was not as substantial as seen in some other models. The range of all the perplexity scores is between 6.18 for the Q8\_0 quantization level and 6.98 for the Q2\_K quantization level, as shown in Table \ref{tab:gemma-2b}. 

When analyzing performance scores, the MMLU-Pro scores remained relatively consistent across most quantization levels. However, the Q2\_K level experienced a drop to 24.29\% from 28.57\% at the Q3\_K level. The MuSR scores showed minimal variation across all quantization levels and remained relatively the same. Interestingly, the CRUXEval score saw a significant increase from the Q4\_K level, with a score of 10.74\% to the highest of 19.60\% for the Q2\_K quantization. Overall, the mean score remained stable throughout all quantization levels.

\section{Model Efficiency}
\begin{figure*}[tb] 
    \centering
    \includegraphics[width=1.0\textwidth]{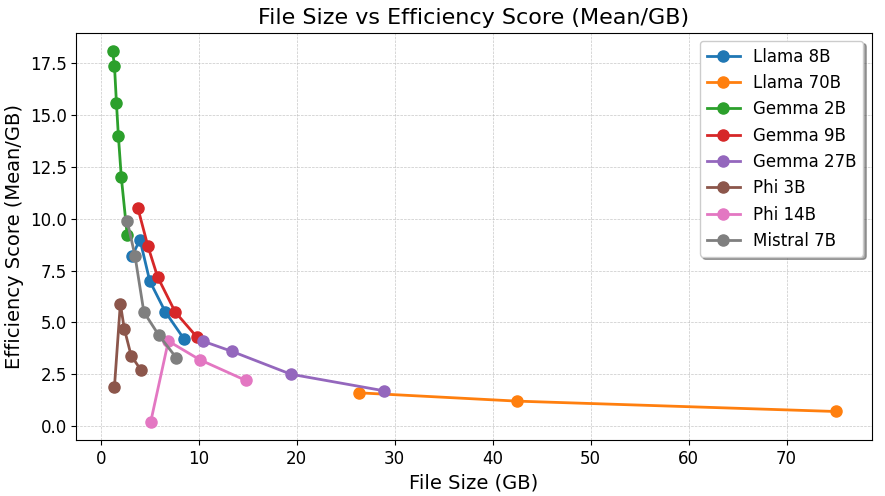}
    \caption{Model scores.}
    \label{fig:efficiency}
\end{figure*}

Figure \ref{fig:efficiency} shows the efficiency score of different language models in relation to their respective file sizes. It reveals that as the size increases, overall the efficiency decreases. The trend suggests that larger models are less efficient. However, the efficiency inherently favors smaller sized models and should be interpreted alongside absolute accuracy scores to fully appreciate the performance-size trade-offs.

\section{Discussion}
Our evaluation confirms that LLM quantization generally degrades performance. This observation broadly aligns with findings on the viability of moderate quantization \citep{jin2024comprehensiveevaluationquantizationstrategies}. However, the extent of the degradation is highly variable and contingent on several factors. While perplexity generally tracks task accuracy, our results also suggest that it is not always a sufficient standalone metric for predicting performance.

We observed significant differences in resilience across model families. For instance, Llama 3 and Gemma 2 maintained performance relatively well even at lower bit precision (e.g., Q3\_K or Q2\_K), whereas Phi 3 was more sensitive to quantization, suffering severe performance loss. This highlights that model architecture plays a critical role in quantization robustness.

Furthermore, the type of task significantly influences the performance of the quantized model. Tasks that require complex reasoning or knowledge, such as MMLU-Pro, tend to be more affected than tasks focused on code comprehension (CRUXEval) or text understanding (MuSR). Low-bit quantization levels risk severe performance loss, particularly for complex, long-text tasks or smaller models.

\section{Conclusion}
This study confirms that while LLM quantization offers significant efficiency benefits, it introduces performance degradation whose severity is highly context-dependent. Our key finding is that the impact of quantization varies substantially based on both the specific model architecture and the nature of the task. Therefore, there is no universally optimal quantization level, and moderate bit precisions (e.g., Q3\_K-Q6\_K) often provide a practical compromise. We conclude that selecting the optimal quantization strategy necessitates a careful, application-specific evaluation considering the chosen model, the task requirements, and the acceptable performance trade-offs.

\section*{Limitations}
The metrics we used to assess the model performances, although widely adopted, may not always correlate with human judgment. This means that the figures and their interpretation should be taken as indicative and not as an ultimate assessment.

In addition, large language models may show a bias and be subject to hallucination. The rankings we observed does not guarantee that the most effective models are not prone to mistakes or misleading answers.

\section*{Acknowledgments}
This work was partially supported by Vetenskaprådet, the Swedish Research Council, registration number 2021-04533.

\bibliography{robin}

\end{document}